\documentclass[a4paper,fleqn]{cas-dc}

\usepackage[numbers]{natbib}
\usepackage[ruled]{algorithm2e}
\usepackage{amsmath,amssymb,amsfonts}
\usepackage{graphicx}
\usepackage{textcomp}
\usepackage{bbm}
\usepackage{threeparttable}
\usepackage{xcolor}
\usepackage{subfigure}
\usepackage{fancyhdr}
\def\tsc#1{\csdef{#1}{\textsc{\lowercase{#1}}\xspace}}
\tsc{WGM}
\tsc{QE}

\begin{document}
\let\WriteBookmarks\relax
\let\WriteBookmarks\relax
\let\printorcid\relax
\def\floatpagepagefraction{1}
\def\textpagefraction{.001}
\def\appendixname{Appendix}
\makeatletter
\renewcommand\appendix{%
  \par%
  \setcounter{section}{0}%
  \setcounter{subsection}{0}%
  \setcounter{equation}{0}%
  \renewcommand\thefigure{\@Alph\c@section.\arabic{figure}}%
  \renewcommand\thetable{\@Alph\c@section.\arabic{table}}%
  \renewcommand\thesection{\appendixname~\@Alph\c@section}%
  \@addtoreset{equation}{section}%
  \renewcommand\theequation{\@Alph\c@section.\arabic{equation}}%
  \addtocontents{toc}{\protect\let\protect\numberline\protect\tmptocnumberline}%
}
\makeatother



\title [mode = title]{Multi-Source Unsupervised Domain Adaptation with Prototype Aggregation}  



%

\author[1]{Min Huang}
\ead{minh@scut.edu.cn}
\cormark[1]
\author[1]{Zifeng Xie}
\ead{202220145216@mail.scut.edu.cn}
\author[2]{Bo Sun}
\ead{sbgz888@163.com}
\author[3]{Ning Wang}
\ead{wangning1@ehv.csg.cn}
\affiliation[1]{organization={South China University of Technology},
            addressline={Guangzhou Higher Education Mega Centre}, 
            city={Guangzhou},
            postcode={510006}, 
            state={Guangdong},
            country={China}}
\affiliation[2]{organization={Institute of International Services Outsourcing},
            addressline={GuangDong University of Foreign Studies}, 
            city={Guangzhou},
            postcode={510006}, 
            state={Guangdong},
            country={China}}
\affiliation[3]{organization={Operation and Maintenance Center of Information and Communication},
            addressline={CSG EHV Power Transmission Company}, 
            city={Guangzhou},
            postcode={510663}, 
            state={Guangdong},
            country={China}}
\cortext[cor1]{Corresponding author.}

















\begin{abstract}
Multi-source domain adaptation (MSDA) plays an important role in industrial model generalization. Recent efforts on MSDA focus on enhancing multi-domain distributional alignment while omitting three issues, e.g., the class-level discrepancy quantification, the unavailability of noisy pseudo-label, and source transferability discrimination, potentially resulting in suboptimal adaption performance. Therefore, we address these issues by proposing a prototype aggregation method that models the discrepancy between source and target domains at the class and domain levels. Our method achieves domain adaptation based on a group of prototypes (i.e., representative feature embeddings). A similarity score-based strategy is designed to quantify the transferability of each domain. At the class level, our method quantifies class-specific cross-domain discrepancy according to reliable target pseudo-labels. At the domain level, our method establishes distributional alignment between noisy pseudo-labeled target samples and the source domain prototypes. Therefore, adaptation at the class and domain levels establishes a complementary mechanism to obtain accurate predictions. The results on three standard benchmarks demonstrate that our method outperforms most state-of-the-art methods. In addition, we provide further elaboration of the proposed method in light of the interpretable results obtained from the analysis experiments.
\end{abstract}


\begin{highlights}
\item Two prototype aggregation discrepancy metrics alleviate the pseudo-label noise.
\item Realizing multi-source domain adaptation in a complementary adaptation mechanism.
\item Quantifying the domain-transferability by a similarity score-based strategy.
\item Experimental and theoretical analyses demonstrate the method’s efficacy.

\end{highlights}



\maketitle


\section{Introduction}\label{intro}
Industrial data is generally characterized as large-scale, originating from various sources, and lacking labeling. Single-source domain adaptation (SDA) is an effective unsupervised learning technique that adapts an industrial model trained on one domain to an unknown domain. 
However, in multi-source scenarios, labeled industrial data originate from various sources and demonstrate diverse characteristics. More training data does not necessarily lead to better adaptation performance \cite{30}. It is illogical to combine all labeled data into a single source and then apply the SDA method to establish distributional alignment.

\begin{figure}[htbp]
\centering
\includegraphics[scale=0.4]{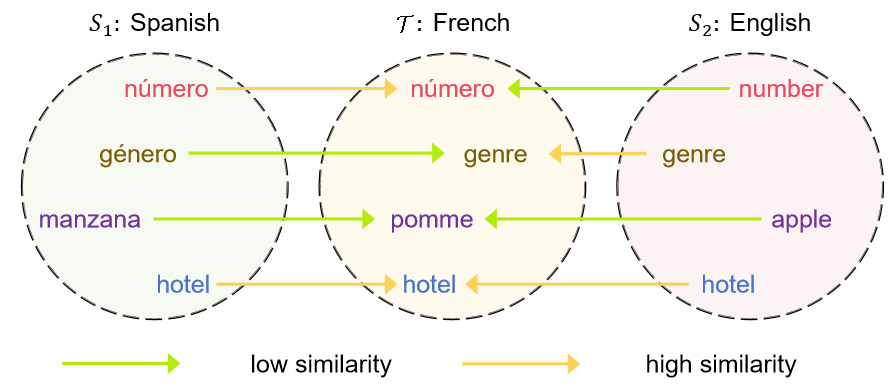}
\vspace{-0.2cm}
\caption{Different transferability of source domains.}
\label{fig:sst}
\end{figure}

After realizing these problems, researchers proposed multi-source domain adaptation (MSDA) to extend domain adaptation to multiple source scenarios. MSDA is developed to enhance multiple distributional alignments and to mitigate various domain shifts \cite{17}. Although many efforts surround the MSDA topic, three primary issues commonly arise. First, most MSDA methods \cite{32,33} focus only on quantifying the domain-level discrepancies while disregarding the class-level discrepancies. Since the class proportions of source and target batches vary, these methods can lead to alignment between samples belonging to different classes. Second, unreliable pseudo-labels have been a long-standing problem for the class-level MSDA methods. On the one hand, incorrect pseudo-labels can lead to negative transfer \cite{107}. On the other hand, excluding the noisy pseudo-labeled target samples \cite{30,100} would hinder the knowledge extraction. Third, many MSDA efforts \cite{31,30} treat each source domain without discrimination, but the common knowledge of each source domain is diverse. As depicted in Fig.\ref{fig:sst}, when French serves as the target domain $\mathcal{T}$, the transferability of the source domain $S_1$ (e.g., Spanish) and the source domain $S_2$ (e.g., English) to target domains varies. 

To tackle these issues, this paper proposes a prototype aggregation method for multi-source domain adaptation (PAMDA). Due to the class disproportion of randomly sampled batches, a group of source prototypes (i.e., representative features) is generated for domain alignment. Since the target label information is agnostic, PAMDA leverages the source supervise knowledge to assign pseudo-labels to target samples. To avoid negative transfer from noisy pseudo-labels, we consider modeling the domain discrepancies at the class and domain levels for diverse confidence pseudo-labeled target samples. 
At the class level, we design a class-prototype aggregation discrepancy metric. PAMDA allocates high weights to the source class prototypes that are similar to the corresponding target features so that these source class prototypes can dominate the feature alignment. Such a design is conducive to transferring source-supervised knowledge that is similar to the target features to the target domain. At the domain level, we design a domain-prototype aggregation discrepancy metric. Similarly, through the weight assignment of PAMDA, the source domain prototypes similar to the noisy pseudo-labeled target features can dominate the feature alignment. Adaptation at the class and domain levels establishes a complementary mechanism where the class-prototype and domain-prototype aggregation discrepancy metrics complement each other. On the one hand, minimizing the class-prototype aggregation discrepancy metric can establish alignment for high-confidence pseudo-labeled target samples, thus facilitating the class-discriminability of the model. On the other hand, minimizing the domain-prototype aggregation discrepancy metric can establish alignment for low-confidence pseudo-labeled target samples, thus facilitating the model to produce more high-confidence pseudo-labels.

The contributions in this work are summed up as:
\begin{itemize}
	\item We propose a prototype aggregation method to address three issues, e.g., the class-level discrepancy quantification, the unavailability of noisy pseudo-label, and source transferability discrimination. Furthermore, we first propose a complementary mechanism where the class-level and domain-level alignment methods can work together. 
	\item We propose a similarity score-based strategy to assess the transferability of source domains. Additionally, we design two prototype aggregation discrepancy metrics to quantify the cross-domain discrepancy at the class and domain levels.
\end{itemize}

\section{Related Work}
\subsection{Single-source domain adaptation}
Over the past decade, SDA has been commended for its notable achievements. Previous SDA efforts can be categorized into two groups: discrepancy-based and adversarial-based. Discrepancy-based efforts model discrepancies between source and target domains, followed by minimizing inter-domain differences. The most widely used discrepancy metrics are maximum mean discrepancy (MMD) \cite{59}, Kullback–Leibler divergence (K-L divergence) \cite{12}, etc. Adversarial-based efforts \cite{22,54} obfuscate the discriminator's discrimination of the domain to which the sample belongs by training the feature extractor.


\subsection{Multi-source domain adaptation}
MSDA expands the scope of domain adaptation algorithms to encompass multi-source scenarios. Initial efforts \cite{67,70} proved a theorem that the target domain can be modeled as a source domain combination with a confirmed upper bound on error. Building upon the theorem, some efforts \cite{84,30} attempt to combine the target-relevant source domains and filter out the irrelevant source samples. Nevertheless, these efforts can neither explore the class-specific semantic information nor make full use of source data. Some multi-model-based methods \cite{31,32} establish the prediction alignment among multiple models. However, it is illogical to expect the models that have learned diverse source knowledge to produce identical predictions. Wang et al. \cite{30} proposed a graph-structured method that forces the feature discrepancy between two arbitrary classes to be consistent across all domains. Two years later, they proposed another graphical method, i.e., Markov Random Field for MSDA (MRF-MSDA). Both graphical methods prioritize aligning reliable pseudo-labeled target samples yet filter out unreliable ones, causing the loss of source information.


\section{Method}
\subsection{Problem description}
In the MSDA scenario, we suppose that there are $N$ labeled source domain $S=\{S_j| j=1, 2, \ldots, N\}$ and one unlabeled target domain $\mathcal{T}$. Each domain exhibits different data characteristics. The $j$-th source domain $S_j=\{(X_{S_j}^{(i)},Y_{S_j}^{(i)})|i=1, 2, \ldots, n_{S_j}\}$ is a collection of $n_{S_j}$ samples, where $Y_{S_j}^{(i)}\in \left\{1,2,\ldots,K \right\}$ ($K$ is the number of classes) is the corresponding classification label of $X_{S_j}^{(i)}$. Simultaneously, the target domain $\mathcal{T}=\{X_{\mathcal{T}}^{(i)}|i=1, 2, \ldots, n_{\mathcal{T}}\}$ is a collection of $n_{\mathcal{T}}$ samples without observable labels. In addition, the PAMDA model is built on two groups of class prototypes, where the source class-prototype group is defined as $\mathbb{B}_s=\left\{b_{S_j}^{ (k))}|j= 1, 2, \ldots, N; k=1, 2, \ldots, K\right\}$, and the target class-prototype group is denoted as $\mathbb{B}_{\mathcal{T}}=\left\{b_{\mathcal{T}}^{ (k)}|k=1, 2, \ldots, K\right\}$.

\begin{figure*}[htbp]
\centering
	\includegraphics[scale=.3]{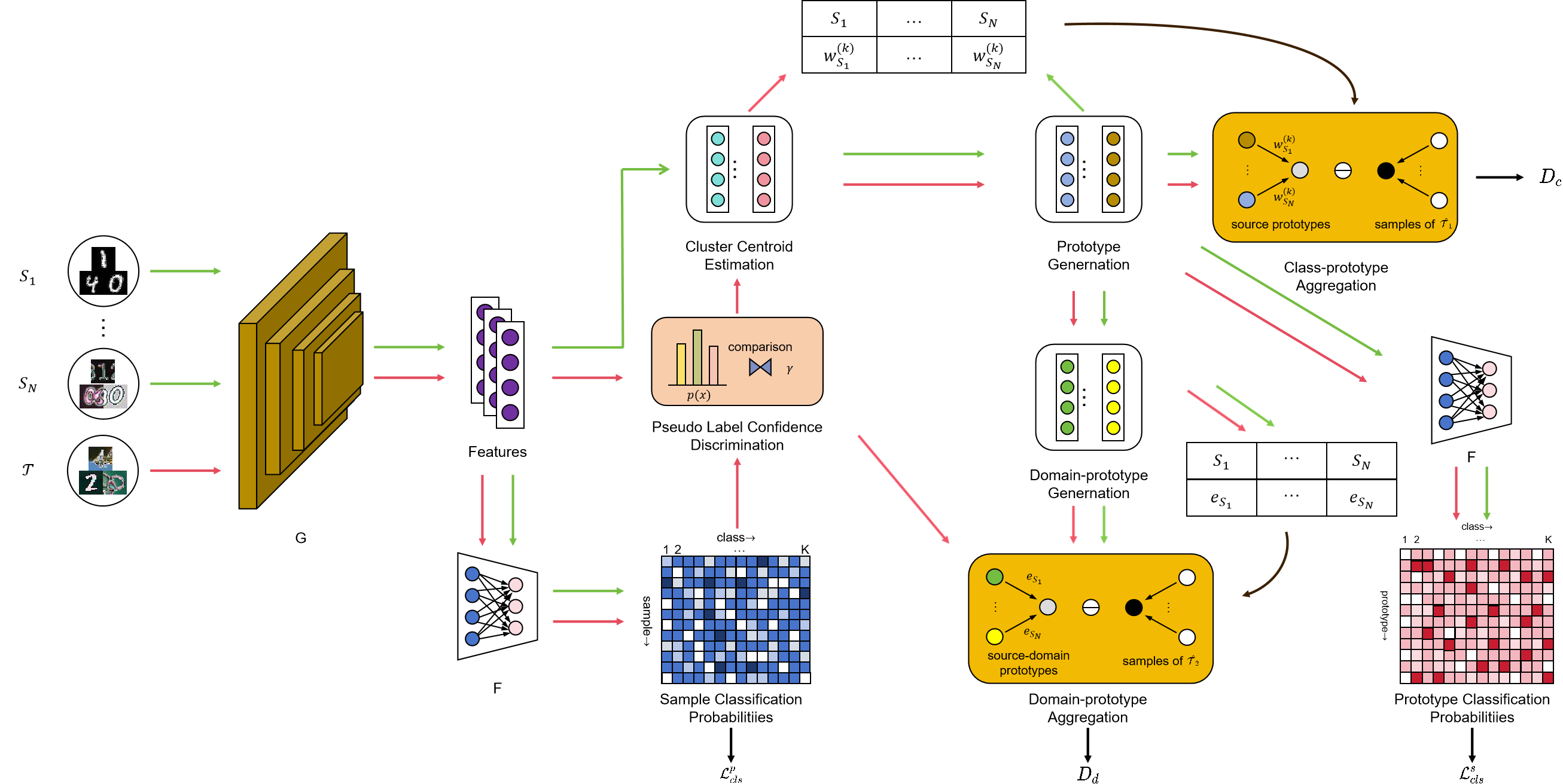}
 
\vspace{-0.2cm}
\caption{Overview of PAMDA. PAMDA is established on a group of prototypes. (a) Each class prototype is generated according to reliable labels. (b) We design two discrepancy metrics for diverse confidence pseudo-labeled target samples. At the class level, a class-prototype aggregation discrepancy is adopted for class alignment across multiple domains. At the domain level, we adopt a domain-prototype aggregation discrepancy for cross-domain alignment based on a group of domain prototypes (i.e., the mean of all class prototypes in the same domain). (c) Last but not least, we design a source classification loss and a prototype classification loss to drive the model to learn the supervised knowledge from source data and class prototypes, respectively.}
\label{fig:123}
\end{figure*}

As illustrated in Fig.\ref{fig:123}, PAMDA involves three stages: prototype generation, prototype aggregation, and objective construction.

\subsection{Prototype generation}\label{pg}
The PAMDA model is constructed to be compatible with mini-batch gradient descent. In each iteration step, a randomly sampled mini-batch $\{\hat{S}_1, \hat{S}_2, \ldots, \hat{S}_N, \hat{\mathcal{T}}\}$ is fed into a feature extractor $G(\cdot)$, which maps the samples into a low-dimensional latent embedding space. The batch size of each domain is denoted as $m$. To describe the class semantic information, the PAMDA model maintains a group of class-specific prototypes for each class. Specifically, the source cluster centroid $\hat{b}^{ (k)}_{S_j}$ is defined as the average of all embeddings belonging to the $k$-th class in $\hat{S}_{j}$.
\begin{align}\label{eq1} 
\hat{b}^{ (k)}_{S_j}=\frac{1}{m}\underset{X_{S_j}\in\hat{S}_{j}^{ (k)}}{\sum}G(X_{S_j}),
\end{align}
where $\hat{S}_{j}^{ (k)}$ represents the cluster of samples belonging to the $k$-th class in $\hat{S}_j$, and $G(\cdot)$ is the feature extraction backbone.

Since the target label information is agnostic, we introduce a confidence-aware scheme for PAMDA to allocate pseudo-labels to target samples. We can obtain the classification probabilities of source and target samples by using $G(\cdot)$ and the classifier $F(\cdot)$. Subsequently, we leverage a logistic-adaptive threshold $\gamma$ proposed in \cite{36} for sample selection. $\gamma$ is defined as 
$\gamma=\frac{1}{1+e^{-3C}}$, where $C$ is the classification accuracy of the source samples.

We can obtain the pseudo-labels of target samples and the corresponding confidence of these pseudo-labels as $\hat{Y}_{\mathcal{T}}^{(i)}=\text{arg}\underset{k}{\text{min}}\quad p(k|X_{\mathcal{T}}^{(i)})$ and $f_{\mathcal{T}}^{(i)}=\underset{k}{\text{min}}\quad p(k|X_{\mathcal{T}}^{(i)})$, respectively, where p$(k|x)$ is the probability that $x$ belongs to $k$-th class.
According to $\gamma$ and $f_{\mathcal{T}}^{(i)}$, $\hat{\mathcal{T}}$ can be divided into a high-confidence subset $\hat{\mathcal{T}}_1$ and a low-confidence subset $\hat{\mathcal{T}}_2$. To avoid negative transfer from noisy pseudo-labels, the target cluster centroid $\hat{b}^{ (k)}_{\mathcal{T}}$ is reformulated as the average of all embeddings belonging to the $k$-th class in $\hat{\mathcal{T}}_1$.
\begin{align}\label{eq6} 
\hat{b}^{ (k)}_{\hat{\mathcal{T}}}=\frac{1}{|\hat{\mathcal{T}}_1^{(k)}|}\underset{X_{\mathcal{T}}\in\hat{\mathcal{T}}_1^{ (k)}}{\sum}G(X_{\mathcal{T}}),
\end{align}
where $\hat{\mathcal{T}}_1^{(k)}$ represents the cluster of samples belonging to the $k$-th class in $\hat{\mathcal{T}}_1$.

The PAMDA model is established on multiple mini-batches of samples. The estimated biases of the cluster centroids in each iteration may vary significantly. Therefore, a momentum update strategy is adopted for prototype generation.
\begin{align}\label{eq7}
b^{ (k)}_{S_j}\leftarrow \eta b^{ (k)}_{S_j} + (1-\eta)\hat{b}^{ (k)}_{S_j},\quad\quad j=1,2,\ldots ,N
\end{align}
\vspace{-0.5cm}
\begin{align}\label{eq8}
b^{ (k)}_{\mathcal{T}}\leftarrow \eta b^{ (k)}_{\mathcal{T}} + (1-\eta)\hat{b}^{ (k)}_{\mathcal{T}},
\end{align}
where $\eta$ is the momentum coefficient constant of 0.7 in all experiments. Analogous strategies have been widely adopted in \cite{30,100} to stabilize the prototype generation process.
\subsection{Prototype aggregation}\label{pa}
Due to the diverse quality of target pseudo-labels, we design two prototype aggregation discrepancies metrics to quantify the discrepancy between source and target domains.

\textbf{(1) Class-prototype aggregation:} As mentioned in Section \ref{intro}, the class features of different source domains significantly differ in transferability. Prioritizing the source class features compatible with the target domain for alignment is more conducive to transferring the source supervisory signal to the target domain. Therefore, we design a class-prototype similarity weight w.r.t. each class $k$.
\begin{align}\label{eq9}
w^{(k)}_{S_j}=\frac{exp( <b^{ (k)}_{S_j}, \hat{b}^{ (k)}_{\mathcal{T}}>/\tau_c)}{\sum_{n=1}^{N} exp( <b^{ (k)}_{S_n}, \hat{b}^{ (k)}_{\mathcal{T}}>/\tau_c)},
\end{align}
where $\tau_c$ is a class hyper-parameter, and $<x,y>=\frac{x^T\cdot y}{||x||\cdot||y||}$.

 Inspired by the maximum mean discrepancy \cite{59}, we design a class-prototype aggregation discrepancy to establish category alignment for each class. The class-prototype aggregation discrepancy $D_c^{(k)}$ is reformulated as:
\begin{align}\label{eq11}
D_c^{(k)}=\left \| \overset{N}{\underset{j=1}{\sum}} w^{(k)}_{S_j}\phi (b^{(k)}_{S_j}) -  \frac{1}{|\hat{\mathcal{T}}_1^{(k)}|}\underset{X_{\mathcal{T}}\in\hat{\mathcal{T}}_1^{ (k)}}{\sum}\phi(G(X_{\mathcal{T}}) )\right \|^2,
\end{align}
where $\phi (\cdot)$ is the gaussian kernel function. The overall class-prototype aggregation discrepancy $D_c$ is denoted as:
\begin{align}\label{eq12}
D_c=\frac{1}{K}\overset{K}{\underset{k=1}{\sum}}D_c^{(k)}.
\end{align}

\textbf{(2) Domain-prototype aggregation:} With unreliable pseudo-labels, the $\hat{\mathcal{T}}_2$ samples can introduce negative transfer to class-level alignment. Therefore, we design domain-prototype aggregation discrepancy for these target samples. The domain-prototype similarity weight w.r.t each source domain is reformulated as:
\begin{align}\label{eq13}
e_{S_j}=\frac{exp( <v_{S_j}, v_{\mathcal{T}}>/\tau_d)}{\sum_{n=1}^{N} exp( <v_{S_n}, v_{\mathcal{T}}>/\tau_d)},
\end{align}
\vspace{-0.5cm}
\begin{align}\label{eq14}
v_{S_j}=\frac{1}{K}\overset{K}{\underset{k=1}{\sum}}b^{(k)}_{S_j},\quad v_{\mathcal{T}}=\frac{1}{|\hat{\mathcal{T}}_2|}\underset{X_{\mathcal{T}}\in\hat{\mathcal{T}}_2}{\sum}G(X_{\mathcal{T}}) .
\end{align}
where $\tau_d$ is a domain hyper-parameter, $v_{S_j}$ is the source-domain prototype of $S_j$, and $v_{\mathcal{T}}$ is the target-domain prototype of $\mathcal{T}$. The domain-prototype aggregation discrepancy $D_d$ is defined as:
\begin{align}\label{eq16}
D_d=\left \| \overset{N}{\underset{j=1}{\sum}} \overset{K}{\underset{k=1}{\sum}}e_{S_j}\phi (b^{(k)}_{S_j}) -  \frac{1}{|\hat{\mathcal{T}}_2|}\underset{X_{\mathcal{T}}\in\hat{\mathcal{T}}_2}{\sum}\phi(G(X_{\mathcal{T}}) )\right \|^2. 
\end{align}

Therefore, the total prototype aggregation discrepancy can be reformulated as follows:
\begin{align}\label{eq17}
D=D_c+D_d
\end{align}
\subsection{Objective construction}\label{oc}
For the PAMDA model training, two optimization objectives are established. One objective is for the PAMDA model to learn supervised information from the source domain effectively. To achieve this, we design a classification loss $\mathcal{L}_{cls}$ to supervise the model training. The other objective is to transfer the supervised knowledge to the target domain. The total prototype aggregation discrepancy is minimized during model training to pursue this objective.

The classification loss $\mathcal{L}_{cls}$ comprises two parts. One is the source classification loss $\mathcal{L}_{cls}^{s}$, which is designed to facilitate source-supervised knowledge learning. Another is the prototype classification loss $\mathcal{L}_{cls}^{p}$. Since the samples used for prototype generation are reliably labeled, the prototypes embody the most representative supervised knowledge. Each prototype is labeled by its corresponding class. We define the classification loss as below:
\begin{align}\label{eq18}
\mathcal{L}_{cls}=\mathcal{L}_{cls}^{s}+\mathcal{L}_{cls}^{p},
\end{align}
\vspace{-0.8cm}
\begin{align}\label{eq19}
\mathcal{L}_{cls}^{s}=\frac{1}{N}\overset{N}{\underset{j=1}{\sum}}\mathbb{E}_{ (X_{S_j},Y_{S_j})\in \hat{S}_j}\mathcal{L}_{ce} (F (G(X_{S_j}),Y_{S_j}),
\end{align}
\vspace{-0.5cm}
\begin{align} \label{eq20}
\begin{split}
\mathcal{L}_{cls}^{p}=\frac{1}{NK}\overset{N}{\underset{j=1}{\sum}}\overset{K}{\underset{k=1}{\sum}}\mathcal{L}_{ce} (F (b^{ (k)}_{S_j}),k)+\frac{1}{K}\overset{K}{\underset{k=1}{\sum}}\mathcal{L}_{ce} (F (b^{ (k)}_{\mathcal{T}}),k),
\end{split}
\end{align} 
where $\mathcal{L}_{ce} (\cdot,\cdot)$ is the cross-entropy loss function.

The overall optimization objectives are as follows: 
\vspace{-0.2cm}
\begin{align} \label{eq21}
\begin{split}
\underset{G}{\text{min}}\quad \mathcal{L}_{cls}+ \alpha D, \quad\quad\quad \underset{F}{\text{min}}\quad \mathcal{L}_{cls}.
\end{split}
\end{align} 

where $\alpha$ is a trade-off parameter. 

\subsection{Theoretical Error Analysis}
Before the experimental verification, we theoretically analyze the target classification errors of the PAMDA algorithm. Suppose $P_S$ and $P_{\mathcal{T}}$ represent the source and target distributions, respectively. Building upon the Theorem 2 proposed by \cite{67}, the target error expectation can be reformulated as:
\begin{align} \label{eq23}
\begin{split}
\sigma_{\mathcal{T}}(h)&=\frac{1}{|\mathcal{T}|}\underset{(X_{\mathcal{T}},Y_{\mathcal{T}})\in \mathcal{T}}{\sum}\mathbbm{1}(Y_{\mathcal{T}},\text{arg}\underset{k}{\text{min}}\quad p(k|X_{\mathcal{T}}))\\
&\leq \sigma_{S}(h)+\frac{1}{2}d_{\mathcal{H}\Delta\mathcal{H}}(P_S,P_{\mathcal{T}})+\beta,
\end{split}
\end{align} 
where $\mathcal{H}$ is a hypothesis space, $h$ ($h \in \mathcal{H}$) is a labeling function (i.e. $G\circ F$), and $Y_{\mathcal{T}}$ is the true label of $X_{\mathcal{T}}$. $\sigma_{S}(h)$ is the source error, which is restricted by the source classification loss. $d_{\mathcal{H}\Delta\mathcal{H}}(P_S,P_{\mathcal{T}})$ is the $\mathcal{H}\Delta\mathcal{H}$-Divergence between the source and target domains, which is restricted by the class-prototype and domain-prototype aggregation discrepancy metrics. $\beta$ is a common error, which is reformulated as $\beta = \underset{ h\in \mathcal{H}}{\text{min}}\quad \epsilon_{\mathcal{T}}(h,f_{\mathcal{T}})+\overset{N}{\underset{j=1}{\sum}}\epsilon_{S_j}(h,f_{S_j})$, where $f_{\mathcal{X}}$ is a true labeling function of domain $\mathcal{X}$ ($\mathcal{X}\in \{\mathcal{T},S_1,S_2, \ldots, S_N\}$), and $\epsilon_{\mathcal{X}}$ is the labeling function discrepancy. $\overset{N}{\underset{j=1}{\sum}}\epsilon_{S_j}(h,f_{S_j})$ is the source classification loss, which can be constrained in our optimization objectives Eqs. \eqref{eq21} and \eqref{eq22}.
Based on \cite{109,67}, $ \epsilon_{\mathcal{T}}(h,f_{\mathcal{T}})$ can be reformulated as:
\begin{align} \label{eq24}
\begin{split}
\epsilon_{\mathcal{T}}(h,f_{\mathcal{T}})\leq \epsilon_{\mathcal{T}}(h,\hat{f}_{\mathcal{T}})+\epsilon_{\mathcal{T}}(f_{\mathcal{T}},\hat{f}_{\mathcal{T}})
\end{split}
\end{align} ,
where $\hat{f}_{\mathcal{T}}$ is a pseudo-labeling function of $\mathcal{T}$. In PAMDA, $\hat{f}_{\mathcal{T}}=h=G\circ F$. Therefore, $\epsilon_{\mathcal{T}}(h,\hat{f}_{\mathcal{T}})=0$. Furthermore, $\epsilon_{\mathcal{T}}(f_{\mathcal{T}},\hat{f}_{\mathcal{T}})$ is constrained by 
the class-prototype and domain-prototype aggregation discrepancy metrics. On the one hand, the class-prototype aggregation discrepancy metric quantifies the discrepancies across domains at the class level. Minimizing the class-prototype aggregation discrepancy metric can enhance the model's class-discriminability for target samples, thus facilitating the PAMDA model to produce more high-quality pseudo-labels. On the other hand, the domain-prototype aggregation discrepancy metric quantifies the discrepancies across domains at the domain level. Minimizing the domain-prototype aggregation discrepancy metric can establish the common domain-feature alignment, facilitating the PAMDA model to transfer source-supervised knowledge to unreliable pseudo-labeled target samples.

Therefore, when the PAMDA is optimized by the objectives Eqs. \eqref{eq21} and \eqref{eq22}, the bound of target error is minimized entirely.
\section{Experiments}

\subsection{Datasets}
\textbf{(1) Digit-5} is a large-scale collection of five digital datasets: MNIST (mt) \cite{75}, MNIST-M (mm) \cite{69}, USPS (up) \cite{65}, SynthDigits (syn) \cite{69}, and SVHN (sv) \cite{66}. We adopt the same data preprocessing program of Ltc-MSDA \cite{30} for our approach.

\textbf{(2) Office\_caltech\_10} \cite{56} is a public benchmark comprising four office-supply image datasets: Amazon (A), Webcam (W), DSLR (D), and Caltech (C). Office\_caltech\_10 includes 9,000 images across ten classes.

\textbf{(3) Office-31} \cite{1} is also an office supply image benchmark, including three sub-datasets: Amazon (A), Webcam (W), and DSLR (D). Office-31 contains a total of 4,110 images across 31 classes.

\subsection{Comparison with other algorithms}
To demonstrate the competitiveness of PAMDA, we introduce the following three experimental setups to these algorithms for multiple-level comparisons. (1) Single Best: We present the best results of the SDA algorithms on all transfer tasks. (2) Source Combination: In this setup, all source domains are combined into one sizeable single-source domain. The transfer tasks are directly accomplished using the SDA algorithms, omitting the domain shifts between the source domains. (3) Multiple Source: Except for PAMDA, the results of the other MSDA algorithms are referenced from the literature where they were originally presented.

For the Single Best and the Source Combination setups, four representative SUDA algorithms (e.g., JAN \cite{76}, MCD \cite{62}, DAN \cite{61}, and ADDA \cite{54}) are selected to compare with PAMDA. In addition, we introduce ten recent MSDA algorithms for the Multiple Source setup. These algorithms are DRT\cite{78}, MDAN \cite{29}, MLAN \cite{31}, $M^3SDA$ \cite{32}, DCTN \cite{74}, MCD \cite{62}, MDDA \cite{33}, Ltc-MSDA \cite{30}, and MRF-MSDA \cite{100}. Source Only is a target-agnostic model trained only with source data.

\subsection{Experimental setups}
All domain adaptation experiments are compatible with mini-batch training and are constructed based on the Pytorch framework. We adopt two fully connected layers as the classifier for all experiments. The 3 conv-2 fc network \cite{30,32}, the AlexNet network \cite{106}, and the ResNet101 network  \cite{11} are employed as the feature extractors on Digits-5, Office-31 \cite{1}, and Office\_caltech\_10 \cite{56}, respectively. The hyper-parameters $\tau_c$ and $\tau_d$ are set to 0.1 and 10, respectively. To mitigate the source noises during the early training, we slowly raise $\alpha$ from 0 to 1 through a transitional strategy\cite{24}: $\alpha = \frac{2}{1+exp(-10t/Max\_Round)}-1$, where $t$ is the index of the current round. 


\subsection{Result}

\textbf{(1) Results on Digits-5}: In Table \ref{tab:II}, we present the comparative experiment results on the digit recognition tasks. Our PAMDA algorithm demonstrates the best mean classification accuracy of 94.2\%, surpassing the state-of-the-art MSDA algorithm MRF-MSDA \cite{100} by a 0.5\% margin. Even on the hard-to-transfer “$\rightarrow mm$” task \cite{78}, the PAMDA algorithm achieves an absolute gain of 4.5\% over the MRF-MSDA algorithm. These competitive results validate the efficacy of the PAMDA algorithm. PAMDA outperforms the state-of-the-art MSDA algorithms for two main reasons. First, PAMDA integrates category-specific knowledge from multiple source domains at a deeper level. Second, the alignment of noisy pseudo-labeled target samples provides more semantic feature information for the PAMDA model.


\begin{table*}[htbp]
\scriptsize
   \centering
   \caption{Comparison in terms of classification accuracy (mean±std\%) on Digits-5}
   \label{tab:II}
   \renewcommand{\arraystretch}{1}
   \begin{threeparttable}
   \resizebox{0.8\textwidth}{!}{
		\begin{tabular}{c c  c c c c c c c}\hline
Standards&Models & $\rightarrow mt$ &  $\rightarrow mm$ & $\rightarrow sy$ & $\rightarrow sv$ & $\rightarrow up$ & Avg \\ 
\hline  &Source Only & 97.2± 0.6& 59.1± 0.6  &  84.6± 0.8 &77.7±0.7&84.7±1.0&80.7\\ 
   
Single Best &ADDA \cite{54} & 97.9±0.8	&71.6±0.5	&86.5±0.6&	75.5±0.5&	92.8±0.7	&	84.8\\ 

&DAN \cite{61} & 96.3±0.5&	63.8±0.7	&85.4±0.8	&62.5±0.7	&94.2±0.9	&80.4\\

\hline &Source Only & 90.2±0.8& 63.4±0.8  &  82.4±0.7&62.9±0.9&88.8±0.8  &77.5 \\

Source Combination &DAN \cite{61} & 97.5±0.6&	67.9±0.8	&86.9±0.5	&67.8±0.6	&93.5±0.8	&	82.7\\ 
&MCD \cite{62} & 96.2±0.8	&72.5±0.7	&87.5±0.7	&78.9±0.8	&95.3±0.7	&	86.1\\

\hline &DCTN \cite{74} & 96.2±0.8&	70.5±1.2	&86.8±0.8	&77.6±0.4	&92.8±0.3	&	84.8\\ 

&DRT \cite{78} & \textbf{99.3±0.1}&	 81.0±0.3	&93.8±0.3	&77.6±0.4	&98.4±0.1	&	91.8\\ 
&$M^3SDA$ \cite{32} & 98.4±0.7&	72.8±1.1	&89.6±0.6	&81.3±0.9	&96.1±0.8	&	87.7\\
Multiple Source &Ltc-MSDA \cite{30} & 99.0±0.4	&85.6±0.8	&93.0±0.5	&83.2±0.6	&98.3±0.4	&	91.8\\

&MLAN \cite{31} &  98.6±0.0&86.3±0.3 & 93.0±0.3 &82.8±0.1& 97.5±0.2 &91.6	\\		&MRF-MSDA \cite{100} &  99.2±0.2&90.7±0.7 & 94.7±0.5 &85.8±0.7& 98.5±0.4 &93.7	 \\
&PAMDA (ours) & 99.1±0.0	&\textbf{95.2±0.3}	&\textbf{95.3±0.2}	&82.7±0.4	&\textbf{98.8±0.1}	&\textbf{94.2}	\\	  
\hline
  \end{tabular}}
  \begin{tablenotes}
      \item[] The best results are marked in bold.
    \end{tablenotes}
  \end{threeparttable}
\end{table*}

\textbf{(2) Results on Office\_caltech\_10}: The result of the PAMDA algorithm and other competitor algorithms are shown in Table \ref{tab:III}. On this dataset, the PAMDA algorithm continues to demonstrate superior performance, achieving a 4.5\%  accuracy improvement compared to the Source Only model. We can observe that the PAMDA algorithm outperforms the state-of-the-art MSDA algorithm $M^3SDA$ \cite{32} by a 1.2\% margin, which consistently demonstrates the effectiveness of the PAMDA algorithm.

\begin{table}[htbp]
\scriptsize
    \small
   \centering
   \caption{Comparison in terms of classification accuracy (mean±std\%) on Office\_caltech\_10}
   \label{tab:III}
   \renewcommand{\arraystretch}{1}
   \setlength{\tabcolsep}{2.5pt}
   \footnotesize
   \begin{threeparttable}
\begin{tabular}{c c  c c c c c }
\hline Models & $\rightarrow W$ &  $\rightarrow D$ & $\rightarrow C$ & $\rightarrow A$  & Avg \\ 
\hline  Source Only & 99.0 &98.3 &87.8 &86.1 &92.8\\ 
 
DCTN \cite{74} & 99.4& 99.0 &90.2& 92.7 &95.3\\

MCD \cite{62} & \textbf{ 99.5}& 99.1 &91.5 &92.1& 95.6 \\
JAN \cite{76} & 99.4 &99.4& 91.2& 91.8 &95.5\\
 $M^3SDA$ \cite{32} & 99.4 &99.2 &91.5& 94.1 &96.1\\ 
PAMDA (ours) & 99.3±0.3	&\textbf{100.0±0.0}	&\textbf{94.6±0.1}	&\textbf{95.2±0.1}	&\textbf{97.3}	\\	  
\hline
  \end{tabular}
  \begin{tablenotes}
      \item[] The best results are marked in bold.
    \end{tablenotes}
  \end{threeparttable}
\end{table}

\textbf{(3) Results on Office-31}: Table \ref{tab:IV} shows that the PAMDA algorithm performs the best classification accuracy on three transfer tasks. Compared to the state-of-the-art algorithm MDDA \cite{33}, the PAMDA algorithm obtains average accuracy improvements of 0.2\%. The favourable results on three datasets demonstrate the stable generalization of the PAMDA algorithm. On Office-31, all MSDA algorithms demonstrated comparable performance on three transfer tasks. We summarize two reasons for these results. First, the performance of MSDA algorithms has reached its limits on the “$\rightarrow D$” and the “$\rightarrow W$” tasks. All MSDA algorithms reported in Table \ref{tab:IV} achieve a classification performance higher than 95\% on these two tasks. Secondly, when the Amazon domain serves as the target domain, two source domains (i.e., the Webcam domain and the DSLR domain) are very similar, indicating little domain gap between the two source domains. This phenomenon significantly limits the advantages of the MSDA algorithms, including our PAMDA algorithm.

\begin{table}[htpb]
\scriptsize
    \small
   \centering
   \caption{Comparison in terms of classification accuracy (mean±std\%) on Office-31}
   \label{tab:IV}
   \renewcommand{\arraystretch}{1}
   \setlength{\tabcolsep}{1pt}
   \footnotesize
   \begin{threeparttable}
		\resizebox{0.45\textwidth}{!}{\begin{tabular}{c c  c c c c c c c}\hline
Standards&Models & $\rightarrow W$ &  $\rightarrow D$ & $\rightarrow A$  & Avg \\  
\hline  &Source Only & 95.3&99.2& 50.3&81.6\\ 
   
Single Best &ADDA \cite{54} & 95.3	&99.4	&54.6	&	83.1\\ 
&DAN \cite{61} & 96.0	&99.0	&54.0	&83.0\\ 

\hline &Source Only & 93.2& 97.7  &  51.6  &80.8 \\
&DAN \cite{61} & 96.2&	98.8	&54.9	&	83.3\\ 
Source Combination &JAN \cite{76} & 95.9	&99.4	&54.6	&	83.3\\
&MCD \cite{62} & 96.2	&99.5	&54.4	&	83.4\\
&ADDA \cite{54} & 96.0	&99.2&	55.9	&	83.7\\

\hline &MDAN \cite{29} &95.4	&99.2	&55.2&	83.3 \\
  
&$M^3SDA$ \cite{32} &96.2	&99.4	&55.4	&	83.7\\
Multiple Source &MDDA \cite{33} & 97.1	&99.2	&56.2	&84.2\\
&DCTN \cite{74} & 96.9&	\textbf{99.6}	&54.9	&	83.8\\
&PAMDA (ours) &\textbf{97.2±0.1} &\textbf{99.6±0.0}	&\textbf{56.5±0.2}	&\textbf{84.4}	\\	  
\hline
  \end{tabular}}
  \begin{tablenotes}
      \item[] The best results are marked in bold.
    \end{tablenotes}
  \end{threeparttable}
\end{table}

\subsection{Ablation analysis}
In this section, we conduct ablation analysis experiments to evaluate the efficacy of each component in our PAMDA model on Digits-5. From Table \ref{tab:V}, we can summarize the following four points. First, class-prototype aggregation can effectively exploit the correlation of class-specific semantic features, thereby establishing a favourable basis for outstanding performance. Second, since the model (3rd row) is only trained with the $\hat{\mathcal{T}}_2$ samples, the domain-prototype aggregation does not provide a significant performance gain for the model. Third, the supervised knowledge of the prototype is favourable to the class discriminability of the model. Finally, the class-prototype aggregation works in conjunction with the domain-prototype aggregation, bringing further performance improvement.

\begin{table}[htpb]
\small
\scriptsize
   \centering
   \caption{Component analysis on Digits-5}

   \label{tab:V}
   \renewcommand{\arraystretch}{1}
   \setlength{\tabcolsep}{2.2pt}
   \footnotesize
\begin{tabular}{c c c c  c c c c c c c c }\hline
$\mathcal{L}_{cls}^s $& $\mathcal{L}_{cls}^p$&$D_c$& $D_d$&\vline  & $\rightarrow mt$ &  $\rightarrow mm$ & $\rightarrow sy$ & $\rightarrow sv$ & $\rightarrow up$ & Avg \\
\hline $\checkmark$&$\checkmark$&&&\vline & 99.1	&65.3	&83.5	&71.3	&96.7	&83.0\\ 
$\checkmark$&$\checkmark$&$\checkmark$& &\vline & 99.1&	94.8	&95.2	&79.0&	98.8	&93.4\\
$\checkmark$&$\checkmark$&& $\checkmark$&\vline & 98.4&	82.2	&88.3	&77.1&	95.3	&88.3\\ 
$\checkmark$&&$\checkmark$&$\checkmark$&\vline & 98.8&	94.3	&94.5	&82.7	&98.2&	93.7\\ 	 
$\checkmark$&$\checkmark$&$\checkmark$&$\checkmark$&\vline &99.1	&95.2	&95.3	&82.7	&98.8&	94.2\\
\hline
  \end{tabular}  
  
\end{table}

\subsection{Hyperparameter analysis} 
\begin{figure}[!htb]
    \centering
    \subfigure[$\tau_c$ ($\tau_d$=10)]{\includegraphics[width=0.49\hsize, height=0.36\hsize]{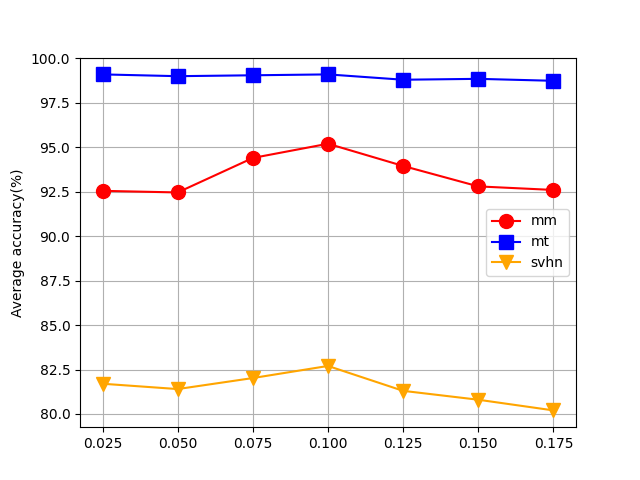}\label{fig:t_c}}\hspace{0.05cm}
    \subfigure[$\tau_d$ ($\tau_c$=0.1)]{\includegraphics[width=0.49\hsize, height=0.36\hsize]{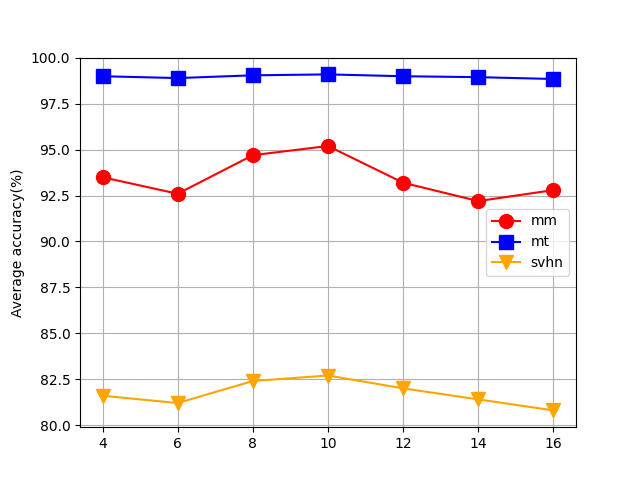}\label{fig:t_d}}
    
    \vspace{-0.2cm}
    \caption{Hyperparameter analysis of $\tau_c$ and $\tau_d$ on Digits-5}
    \label{fig:sen}
\end{figure}

This part presents an experimental analysis of the hyperparameters  $\tau_c$ (i.e., hyperparameter $\tau_c$ in Eq. \eqref{eq9}) and $\tau_d$ (i.e., hyperparameter $\tau_d$ in Eq. \eqref{eq13}). As presented in Fig.\ref{fig:t_c}, when $\tau_d$ is fixed as 10, the optimal performance is achieved at 0.1. As shown in Fig.\ref{fig:t_d}, when $\tau_c$ is fixed as 0.1 and $\tau_d$ is set to 10, the experiment performance reaches the peak value. These results show that PAMDA is not sensitive to $\tau_c$ and $\tau_d$, which substantiates the resilience of our PAMDA algorithm.
\vspace{-0.2cm}
\subsection{Visualization}
\begin{figure}[!htb]
    \centering
    \subfigure[Source Only]{\includegraphics[width=0.46\hsize, height=0.45\hsize]{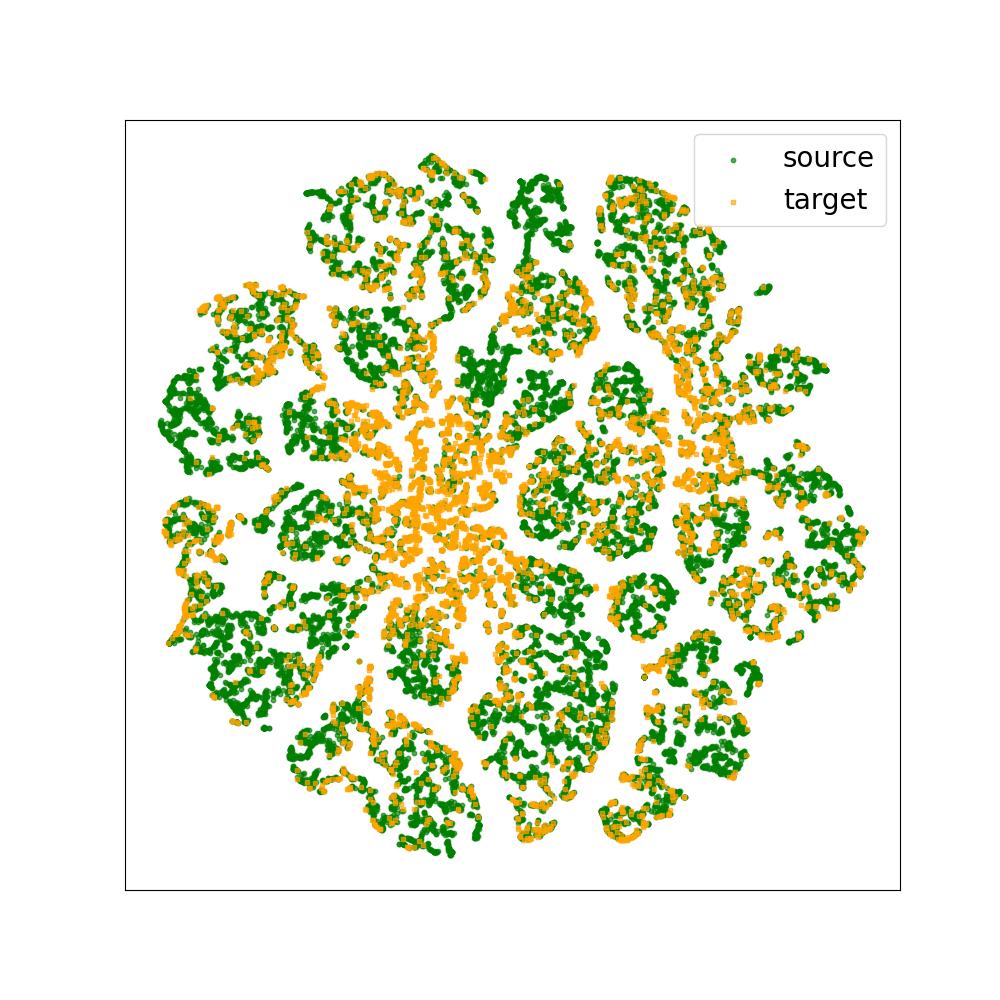}\label{fig:sub1}}\hspace{0.5cm}
    \subfigure[PAMDA]{\includegraphics[width=0.46\hsize, height=0.45\hsize]{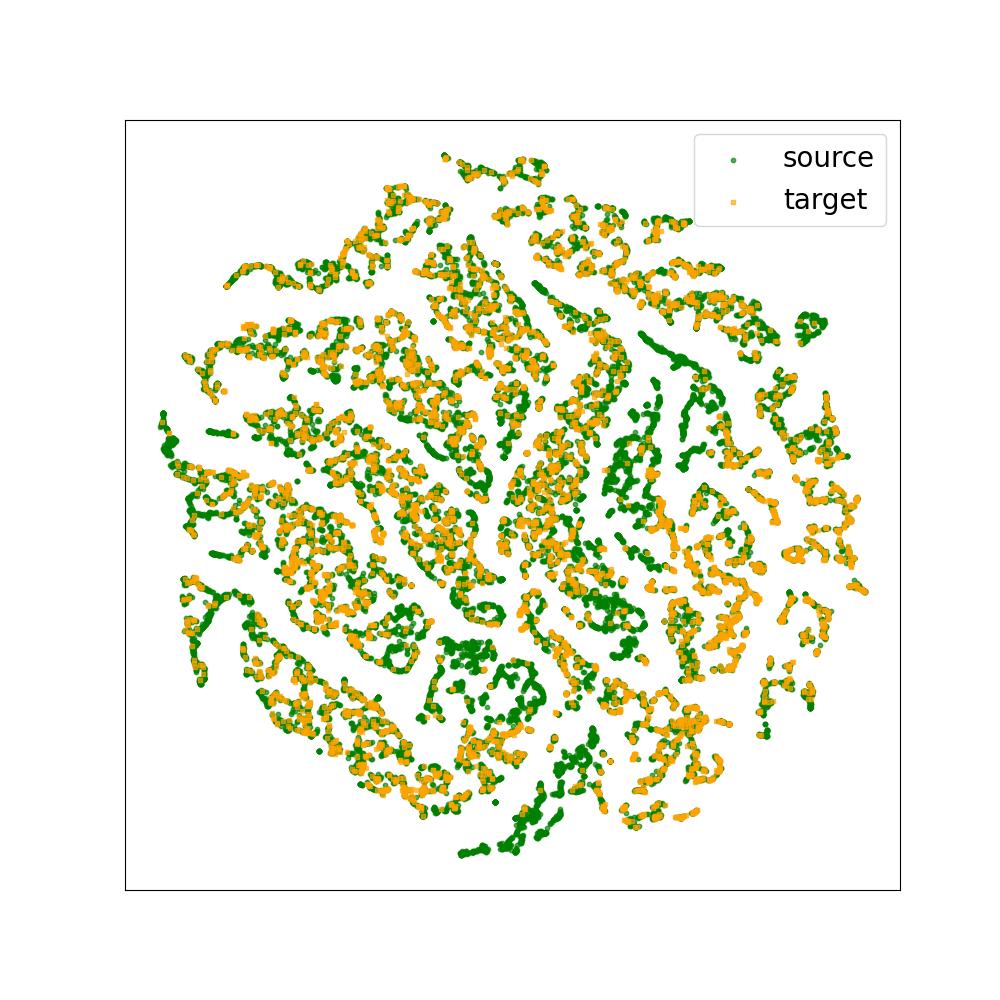}\label{fig:sub2}}
    \vspace{-0.3cm}
    \caption{Feature distributions on the “$\rightarrow mm$” task of Digits-5.}
    \label{fig:whole}
\end{figure}

\textbf{(1) Feature Visualization:} 
Fig.\ref{fig:whole} depicts the feature distributions of Source Only and PAMDA on the “$\rightarrow mm$” task of Digits-5. We project the feature embeddings into a two-dimensional latent space by t-SNE \cite{102} and then visualize the feature space. As shown in Fig.\ref{fig:sub1} (i.e., the scatter plot of Source Only), we can observe the cross-domain distributional discrepancy. In Fig.\ref{fig:sub2} (i.e., the scatter plot of PAMDA), it can observed that the target features are better aligned with source features. These results suggest that our PAMDA model demonstrates greater discriminability for target features compared to the Source Only model, substantiating the result presented in Table \ref{tab:II}.

\begin{figure}[htbp]
\centering
\includegraphics[scale=0.4]{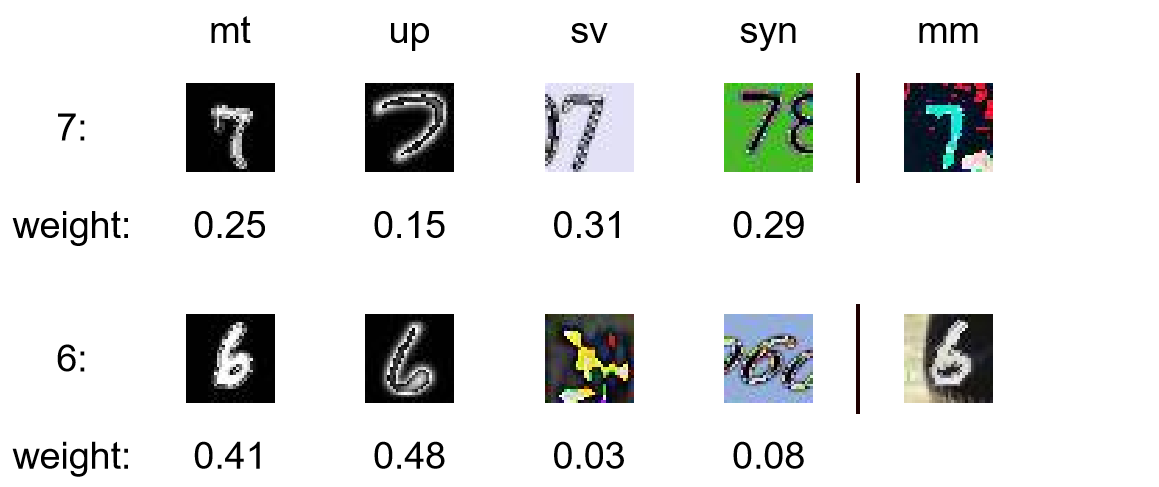}
\vspace{-0.1cm}
\caption{Class weight distributions on the “$\rightarrow mm$” task of Digits-5.}
\label{fig:weight}
\end{figure}

\textbf{(2) Visualization of weight distribution}: To validate the efficacy of the similarity score-based strategy, we visualize the class weight distributions on the “$\rightarrow mm$” task of Digits-5. For clarity, we only display two classes, i.e., 7 and 6. As illustrated in Fig.\ref {fig:weight}, our PAMDA model assigns high weights to the source prototypes that exhibit structural similarities to the corresponding target prototypes. For example, the handwriting style of the number 6 on MNIST-M is more similar to that on MNIST and USPS, while it differs more from the handwriting style on SVHN and SynthDigits. The prototypes of the number 6 on MNIST and USPS are allocated high weights, while those of the number 6 on SVHN and SynthDigits are allocated low weights. These results demonstrate that our similarity score-based strategy is structurally selective for source prototypes.

\section{Conclusion}
Previous MSDA efforts suffer from three issues: class-level discrepancy quantification, the unavailability of noisy pseudo-labels, and source transferability discrimination.
This paper proposes a prototype aggregation method (i.e., PAMDA) for these issues.
The PAMDA algorithm is established on a group of prototypes. Since noisy pseudo-labels are inevitable, we quantify the multiple domain discrepancies at the class and domain levels for diverse confidence pseudo-labeled target samples. Specifically, we propose a similarity score-based strategy to assess the source transferability at the class and domain levels. According to the weight produced by the strategy, we design two prototype aggregation discrepancy metrics for domain discrepancy quantification. As the prototype aggregation discrepancies are continuously minimized, the discriminability of the PAMDA model for target features is constantly improved. The PAMDA achieves 94.2\%, 97.3\%, and 84.4\% average accuracy on three popular public datasets, Digits-5, Office\_caltech\_10, and Office-31, respectively. In addition, further experiments demonstrate that the PAMDA model is competitively robust and exhibits stable generalization. 
In many industrial scenarios, each domain may contain different categories. The sample proportions from each class may vary significantly. Developing our algorithm to address these issues is our future work.

\section*{Declaration of competing interest}
The author declares that the research was conducted without any commercial or financial relationships that could be considered a potential conflict of interest.
\section*{Data availability}
The original contributions presented in the study are included in the article. Further inquiries can be directed to the corresponding author.
\section*{Acknowledgment}
The work described in this paper was funded by the Natural Science Foundation of Guangdong Province (Grant No. 2022A1515011370).
\printcredits
\bibliography{reference}

\begin{thebibliography}{32}
\expandafter\ifx\csname natexlab\endcsname\relax\def\natexlab#1{#1}\fi
\providecommand{\url}[1]{\texttt{#1}}
\providecommand{\href}[2]{#2}
\providecommand{\path}[1]{#1}
\providecommand{\DOIprefix}{doi:}
\providecommand{\ArXivprefix}{arXiv:}
\providecommand{\URLprefix}{URL: }
\providecommand{\Pubmedprefix}{pmid:}
\providecommand{\doi}[1]{\href{http://dx.doi.org/#1}{\path{#1}}}
\providecommand{\Pubmed}[1]{\href{pmid:#1}{\path{#1}}}
\providecommand{\bibinfo}[2]{#2}
\ifx\xfnm\relax \def\xfnm[#1]{\unskip,\space#1}\fi
\bibitem[{Ben-David et~al.(2010)Ben-David, Blitzer, Crammer, Kulesza, Pereira and Vaughan}]{67}
\bibinfo{author}{Ben-David, S.}, \bibinfo{author}{Blitzer, J.}, \bibinfo{author}{Crammer, K.}, \bibinfo{author}{Kulesza, A.}, \bibinfo{author}{Pereira, F.}, \bibinfo{author}{Vaughan, J.W.}, \bibinfo{year}{2010}.
\newblock \bibinfo{title}{A theory of learning from different domains}.
\newblock \bibinfo{journal}{Machine learning} \bibinfo{volume}{79}, \bibinfo{pages}{151--175}.
\bibitem[{Ganin and Lempitsky(2015)}]{24}
\bibinfo{author}{Ganin, Y.}, \bibinfo{author}{Lempitsky, V.}, \bibinfo{year}{2015}.
\newblock \bibinfo{title}{Unsupervised domain adaptation by backpropagation}, in: \bibinfo{booktitle}{International conference on machine learning}, \bibinfo{organization}{PMLR}. pp. \bibinfo{pages}{1180--1189}.
\bibitem[{Ganin et~al.(2016)Ganin, Ustinova, Ajakan, Germain, Larochelle, Laviolette, Marchand and Lempitsky}]{69}
\bibinfo{author}{Ganin, Y.}, \bibinfo{author}{Ustinova, E.}, \bibinfo{author}{Ajakan, H.}, \bibinfo{author}{Germain, P.}, \bibinfo{author}{Larochelle, H.}, \bibinfo{author}{Laviolette, F.}, \bibinfo{author}{Marchand, M.}, \bibinfo{author}{Lempitsky, V.}, \bibinfo{year}{2016}.
\newblock \bibinfo{title}{Domain-adversarial training of neural networks}.
\newblock \bibinfo{journal}{The journal of machine learning research} \bibinfo{volume}{17}, \bibinfo{pages}{2096--2030}.
\bibitem[{Gong et~al.(2012)Gong, Shi, Sha and Grauman}]{56}
\bibinfo{author}{Gong, B.}, \bibinfo{author}{Shi, Y.}, \bibinfo{author}{Sha, F.}, \bibinfo{author}{Grauman, K.}, \bibinfo{year}{2012}.
\newblock \bibinfo{title}{Geodesic flow kernel for unsupervised domain adaptation}, in: \bibinfo{booktitle}{2012 IEEE conference on computer vision and pattern recognition}, \bibinfo{organization}{IEEE}. pp. \bibinfo{pages}{2066--2073}.
\bibitem[{Gretton et~al.(2012)Gretton, Borgwardt, Rasch, Sch{\"o}lkopf and Smola}]{59}
\bibinfo{author}{Gretton, A.}, \bibinfo{author}{Borgwardt, K.M.}, \bibinfo{author}{Rasch, M.J.}, \bibinfo{author}{Sch{\"o}lkopf, B.}, \bibinfo{author}{Smola, A.}, \bibinfo{year}{2012}.
\newblock \bibinfo{title}{A kernel two-sample test}.
\newblock \bibinfo{journal}{The Journal of Machine Learning Research} \bibinfo{volume}{13}, \bibinfo{pages}{723--773}.
\bibitem[{He et~al.(2016)He, Zhang, Ren and Sun}]{11}
\bibinfo{author}{He, K.}, \bibinfo{author}{Zhang, X.}, \bibinfo{author}{Ren, S.}, \bibinfo{author}{Sun, J.}, \bibinfo{year}{2016}.
\newblock \bibinfo{title}{Deep residual learning for image recognition}, in: \bibinfo{booktitle}{Proceedings of the IEEE conference on computer vision and pattern recognition}, pp. \bibinfo{pages}{770--778}.
\bibitem[{Hull(1994)}]{65}
\bibinfo{author}{Hull, J.J.}, \bibinfo{year}{1994}.
\newblock \bibinfo{title}{A database for handwritten text recognition research}.
\newblock \bibinfo{journal}{IEEE Transactions on pattern analysis and machine intelligence} \bibinfo{volume}{16}, \bibinfo{pages}{550--554}.
\bibitem[{Krizhevsky et~al.(2017)Krizhevsky, Sutskever and Hinton}]{106}
\bibinfo{author}{Krizhevsky, A.}, \bibinfo{author}{Sutskever, I.}, \bibinfo{author}{Hinton, G.E.}, \bibinfo{year}{2017}.
\newblock \bibinfo{title}{Imagenet classification with deep convolutional neural networks}.
\newblock \bibinfo{journal}{Communications of the ACM} \bibinfo{volume}{60}, \bibinfo{pages}{84--90}.
\bibitem[{LeCun et~al.(1998)LeCun, Bottou, Bengio and Haffner}]{75}
\bibinfo{author}{LeCun, Y.}, \bibinfo{author}{Bottou, L.}, \bibinfo{author}{Bengio, Y.}, \bibinfo{author}{Haffner, P.}, \bibinfo{year}{1998}.
\newblock \bibinfo{title}{Gradient-based learning applied to document recognition}.
\newblock \bibinfo{journal}{Proceedings of the IEEE} \bibinfo{volume}{86}, \bibinfo{pages}{2278--2324}.
\bibitem[{Li et~al.(2021)Li, Yuan, Chen, Wang and Vasconcelos}]{78}
\bibinfo{author}{Li, Y.}, \bibinfo{author}{Yuan, L.}, \bibinfo{author}{Chen, Y.}, \bibinfo{author}{Wang, P.}, \bibinfo{author}{Vasconcelos, N.}, \bibinfo{year}{2021}.
\newblock \bibinfo{title}{Dynamic transfer for multi-source domain adaptation}, in: \bibinfo{booktitle}{Proceedings of the IEEE/CVF Conference on Computer Vision and Pattern Recognition}, pp. \bibinfo{pages}{10998--11007}.
\bibitem[{Long et~al.(2015)Long, Cao, Wang and Jordan}]{61}
\bibinfo{author}{Long, M.}, \bibinfo{author}{Cao, Y.}, \bibinfo{author}{Wang, J.}, \bibinfo{author}{Jordan, M.}, \bibinfo{year}{2015}.
\newblock \bibinfo{title}{Learning transferable features with deep adaptation networks}, in: \bibinfo{booktitle}{International conference on machine learning}, \bibinfo{organization}{PMLR}. pp. \bibinfo{pages}{97--105}.
\bibitem[{Long et~al.(2018)Long, Cao, Wang and Jordan}]{22}
\bibinfo{author}{Long, M.}, \bibinfo{author}{Cao, Z.}, \bibinfo{author}{Wang, J.}, \bibinfo{author}{Jordan, M.I.}, \bibinfo{year}{2018}.
\newblock \bibinfo{title}{Conditional adversarial domain adaptation}, in: \bibinfo{booktitle}{Proceedings of the 32nd International Conference on Neural Information Processing Systems}, pp. \bibinfo{pages}{1647--1657}.
\bibitem[{Long et~al.(2017)Long, Zhu, Wang and Jordan}]{76}
\bibinfo{author}{Long, M.}, \bibinfo{author}{Zhu, H.}, \bibinfo{author}{Wang, J.}, \bibinfo{author}{Jordan, M.I.}, \bibinfo{year}{2017}.
\newblock \bibinfo{title}{Deep transfer learning with joint adaptation networks}, in: \bibinfo{booktitle}{International conference on machine learning}, \bibinfo{organization}{PMLR}. pp. \bibinfo{pages}{2208--2217}.
\bibitem[{Van~der Maaten and Hinton(2008)}]{102}
\bibinfo{author}{Van~der Maaten, L.}, \bibinfo{author}{Hinton, G.}, \bibinfo{year}{2008}.
\newblock \bibinfo{title}{Visualizing data using t-sne.}
\newblock \bibinfo{journal}{Journal of machine learning research} \bibinfo{volume}{9}.
\bibitem[{Mansour et~al.(2008)Mansour, Mohri and Rostamizadeh}]{70}
\bibinfo{author}{Mansour, Y.}, \bibinfo{author}{Mohri, M.}, \bibinfo{author}{Rostamizadeh, A.}, \bibinfo{year}{2008}.
\newblock \bibinfo{title}{Domain adaptation with multiple sources}.
\newblock \bibinfo{journal}{Advances in neural information processing systems} \bibinfo{volume}{21}.
\bibitem[{Pan and Yang(2009)}]{107}
\bibinfo{author}{Pan, S.J.}, \bibinfo{author}{Yang, Q.}, \bibinfo{year}{2009}.
\newblock \bibinfo{title}{A survey on transfer learning}.
\newblock \bibinfo{journal}{IEEE Transactions on knowledge and data engineering} \bibinfo{volume}{22}, \bibinfo{pages}{1345--1359}.
\bibitem[{Peng et~al.(2019)Peng, Bai, Xia, Huang, Saenko and Wang}]{32}
\bibinfo{author}{Peng, X.}, \bibinfo{author}{Bai, Q.}, \bibinfo{author}{Xia, X.}, \bibinfo{author}{Huang, Z.}, \bibinfo{author}{Saenko, K.}, \bibinfo{author}{Wang, B.}, \bibinfo{year}{2019}.
\newblock \bibinfo{title}{Moment matching for multi-source domain adaptation}, in: \bibinfo{booktitle}{Proceedings of the IEEE/CVF international conference on computer vision}, pp. \bibinfo{pages}{1406--1415}.
\bibitem[{Saenko et~al.(2010)Saenko, Kulis, Fritz and Darrell}]{1}
\bibinfo{author}{Saenko, K.}, \bibinfo{author}{Kulis, B.}, \bibinfo{author}{Fritz, M.}, \bibinfo{author}{Darrell, T.}, \bibinfo{year}{2010}.
\newblock \bibinfo{title}{Adapting visual category models to new domains}, in: \bibinfo{booktitle}{Computer Vision--ECCV 2010: 11th European Conference on Computer Vision, Heraklion, Crete, Greece, September 5-11, 2010, Proceedings, Part IV 11}, \bibinfo{organization}{Springer}. pp. \bibinfo{pages}{213--226}.
\bibitem[{Saito et~al.(2018)Saito, Watanabe, Ushiku and Harada}]{62}
\bibinfo{author}{Saito, K.}, \bibinfo{author}{Watanabe, K.}, \bibinfo{author}{Ushiku, Y.}, \bibinfo{author}{Harada, T.}, \bibinfo{year}{2018}.
\newblock \bibinfo{title}{Maximum classifier discrepancy for unsupervised domain adaptation}, in: \bibinfo{booktitle}{Proceedings of the IEEE conference on computer vision and pattern recognition}, pp. \bibinfo{pages}{3723--3732}.
\bibitem[{T{\'o}th and Gosztolya(2016)}]{12}
\bibinfo{author}{T{\'o}th, L.}, \bibinfo{author}{Gosztolya, G.}, \bibinfo{year}{2016}.
\newblock \bibinfo{title}{Adaptation of dnn acoustic models using kl-divergence regularization and multi-task training}, in: \bibinfo{booktitle}{Speech and Computer: 18th International Conference, SPECOM 2016, Budapest, Hungary, August 23-27, 2016, Proceedings 18}, \bibinfo{organization}{Springer}. pp. \bibinfo{pages}{108--115}.
\bibitem[{Tzeng et~al.(2017)Tzeng, Hoffman, Saenko and Darrell}]{54}
\bibinfo{author}{Tzeng, E.}, \bibinfo{author}{Hoffman, J.}, \bibinfo{author}{Saenko, K.}, \bibinfo{author}{Darrell, T.}, \bibinfo{year}{2017}.
\newblock \bibinfo{title}{Adversarial discriminative domain adaptation}, in: \bibinfo{booktitle}{Proceedings of the IEEE conference on computer vision and pattern recognition}, pp. \bibinfo{pages}{7167--7176}.
\bibitem[{Wang et~al.(2020)Wang, Xu, Ni and Zhang}]{30}
\bibinfo{author}{Wang, H.}, \bibinfo{author}{Xu, M.}, \bibinfo{author}{Ni, B.}, \bibinfo{author}{Zhang, W.}, \bibinfo{year}{2020}.
\newblock \bibinfo{title}{Learning to combine: Knowledge aggregation for multi-source domain adaptation}, in: \bibinfo{booktitle}{Computer Vision--ECCV 2020: 16th European Conference, Glasgow, UK, August 23--28, 2020, Proceedings, Part VIII 16}, \bibinfo{organization}{Springer}. pp. \bibinfo{pages}{727--744}.
\bibitem[{Wen et~al.(2020)Wen, Greiner and Schuurmans}]{84}
\bibinfo{author}{Wen, J.}, \bibinfo{author}{Greiner, R.}, \bibinfo{author}{Schuurmans, D.}, \bibinfo{year}{2020}.
\newblock \bibinfo{title}{Domain aggregation networks for multi-source domain adaptation}, in: \bibinfo{booktitle}{International conference on machine learning}, \bibinfo{organization}{PMLR}. pp. \bibinfo{pages}{10214--10224}.
\bibitem[{Xu et~al.(2022a)Xu, Wang and Ni}]{100}
\bibinfo{author}{Xu, M.}, \bibinfo{author}{Wang, H.}, \bibinfo{author}{Ni, B.}, \bibinfo{year}{2022}a.
\newblock \bibinfo{title}{Graphical modeling for multi-source domain adaptation}.
\newblock \bibinfo{journal}{IEEE Transactions on Pattern Analysis and Machine Intelligence} .
\bibitem[{Xu et~al.(2018)Xu, Chen, Zuo, Yan and Lin}]{74}
\bibinfo{author}{Xu, R.}, \bibinfo{author}{Chen, Z.}, \bibinfo{author}{Zuo, W.}, \bibinfo{author}{Yan, J.}, \bibinfo{author}{Lin, L.}, \bibinfo{year}{2018}.
\newblock \bibinfo{title}{Deep cocktail network: Multi-source unsupervised domain adaptation with category shift}, in: \bibinfo{booktitle}{Proceedings of the IEEE conference on computer vision and pattern recognition}, pp. \bibinfo{pages}{3964--3973}.
\bibitem[{Xu et~al.(2022b)Xu, Kan, Shan and Chen}]{31}
\bibinfo{author}{Xu, Y.}, \bibinfo{author}{Kan, M.}, \bibinfo{author}{Shan, S.}, \bibinfo{author}{Chen, X.}, \bibinfo{year}{2022}b.
\newblock \bibinfo{title}{Mutual learning of joint and separate domain alignments for multi-source domain adaptation}, in: \bibinfo{booktitle}{Proceedings of the IEEE/CVF Winter Conference on Applications of Computer Vision}, pp. \bibinfo{pages}{1890--1899}.
\bibitem[{Yosinski et~al.(2014)Yosinski, Clune, Bengio and Lipson}]{17}
\bibinfo{author}{Yosinski, J.}, \bibinfo{author}{Clune, J.}, \bibinfo{author}{Bengio, Y.}, \bibinfo{author}{Lipson, H.}, \bibinfo{year}{2014}.
\newblock \bibinfo{title}{How transferable are features in deep neural networks?}
\newblock \bibinfo{journal}{Advances in neural information processing systems} \bibinfo{volume}{27}, \bibinfo{pages}{3320--3328}.
\bibitem[{Yuval(2011)}]{66}
\bibinfo{author}{Yuval, N.}, \bibinfo{year}{2011}.
\newblock \bibinfo{title}{Reading digits in natural images with unsupervised feature learning}, in: \bibinfo{booktitle}{Proceedings of the NIPS Workshop on Deep Learning and Unsupervised Feature Learning}.
\bibitem[{Zhang et~al.(2018)Zhang, Ouyang, Li and Xu}]{36}
\bibinfo{author}{Zhang, W.}, \bibinfo{author}{Ouyang, W.}, \bibinfo{author}{Li, W.}, \bibinfo{author}{Xu, D.}, \bibinfo{year}{2018}.
\newblock \bibinfo{title}{Collaborative and adversarial network for unsupervised domain adaptation}, in: \bibinfo{booktitle}{Proceedings of the IEEE conference on computer vision and pattern recognition}, pp. \bibinfo{pages}{3801--3809}.
\bibitem[{Zhao et~al.(2018)Zhao, Zhang, Wu, Moura, Costeira and Gordon}]{29}
\bibinfo{author}{Zhao, H.}, \bibinfo{author}{Zhang, S.}, \bibinfo{author}{Wu, G.}, \bibinfo{author}{Moura, J.M.}, \bibinfo{author}{Costeira, J.P.}, \bibinfo{author}{Gordon, G.J.}, \bibinfo{year}{2018}.
\newblock \bibinfo{title}{Adversarial multiple source domain adaptation}.
\newblock \bibinfo{journal}{Advances in neural information processing systems} \bibinfo{volume}{31}.
\bibitem[{Zhao et~al.(2020)Zhao, Wang, Zhang, Gu, Li, Song, Xu, Hu, Chai and Keutzer}]{33}
\bibinfo{author}{Zhao, S.}, \bibinfo{author}{Wang, G.}, \bibinfo{author}{Zhang, S.}, \bibinfo{author}{Gu, Y.}, \bibinfo{author}{Li, Y.}, \bibinfo{author}{Song, Z.}, \bibinfo{author}{Xu, P.}, \bibinfo{author}{Hu, R.}, \bibinfo{author}{Chai, H.}, \bibinfo{author}{Keutzer, K.}, \bibinfo{year}{2020}.
\newblock \bibinfo{title}{Multi-source distilling domain adaptation}, in: \bibinfo{booktitle}{Proceedings of the AAAI Conference on Artificial Intelligence}, pp. \bibinfo{pages}{12975--12983}.
\bibitem[{Zhou et~al.(2024)Zhou, Li, Ye, Zhu and Tang}]{109}
\bibinfo{author}{Zhou, L.}, \bibinfo{author}{Li, N.}, \bibinfo{author}{Ye, M.}, \bibinfo{author}{Zhu, X.}, \bibinfo{author}{Tang, S.}, \bibinfo{year}{2024}.
\newblock \bibinfo{title}{Source-free domain adaptation with class prototype discovery}.
\newblock \bibinfo{journal}{Pattern recognition} \bibinfo{volume}{145}, \bibinfo{pages}{109974}.

\end{thebibliography}
\small
\bibliographystyle{cas-model2-names}

\end{document}